\documentclass{article}

\usepackage[preprint]{neurips_2025}

\usepackage[utf8]{inputenc} 
\usepackage[T1]{fontenc}    
\usepackage{hyperref}       
\usepackage{url}            
\usepackage{booktabs}       
\usepackage{amsfonts}       
\usepackage{nicefrac}       
\usepackage{microtype}      
\usepackage{xcolor}         
\usepackage{multirow}
\usepackage{bm}
\usepackage{subcaption} 

\usepackage{numprint}
\npthousandsep{\,}

\usepackage{fontawesome5} 
\usepackage{diagbox} 

\usepackage[pdftex]{graphicx}

\usepackage{tikz}
\usetikzlibrary{calc,decorations.pathreplacing,calligraphy}
\newcommand{\tikzmark}[1]{%
  \tikz[overlay,remember picture,baseline] \node [anchor=base] (#1) {};}
\newcommand{\DrawVerticalBrace}[3][]{%
    \begin{tikzpicture}[overlay,remember picture]
        \draw[decorate,decoration={calligraphic brace,mirror,amplitude=1ex}, #1] 
            ($(#3)+(0.0em,-0.5ex)$) to
            ($(#2)+(0.0em,+1.7ex)$);%
    \end{tikzpicture}%
}

\title{Explaining How Visual, Textual and Multimodal Encoders Share Concepts}

\author{%
  Clément Cornet\\
    Université Paris-Saclay, CEA, List,\\
    F-91120, Palaiseau, France  \\
  \texttt{clement.cornet@cea.fr} \\
   \And
   Romaric Besançon \\
   Université Paris-Saclay, CEA, List,\\
   F-91120, Palaiseau, France  \\
   \texttt{romaric.besancon@cea.fr} \\
   \AND
   Hervé Le Borgne \\
   Université Paris-Saclay, CEA, List,\\
   F-91120, Palaiseau, France  \\
   \texttt{herve.le-borgne@cea.fr} \\
}

\begin{document}

\maketitle

\begin{abstract}
Sparse autoencoders (SAEs) have emerged as a powerful technique for extracting human-interpretable features from neural networks activations. Previous works compared different models based on SAE-derived features but those comparisons have been restricted to models within the same modality. We propose a novel indicator allowing quantitative comparison of models across SAE features, and use it to conduct a comparative study of visual, textual and multimodal encoders. We also propose to quantify the \textit{Comparative Sharedness} of individual features between different classes of models. With these two new tools, we conduct several studies on 21 encoders of the three types, with two significantly different sizes, and considering generalist and domain specific datasets. The results allow to revisit previous studies at the light of encoders trained in a multimodal context and to quantify to which extent all these models share some representations or features. They also suggest that visual features that are specific to VLMs among vision encoders are shared with text encoders, highlighting the impact of text pretraining. The code is available at \href{https://github.com/CEA-LIST/SAEshareConcepts}{https://github.com/CEA-LIST/SAEshareConcepts}
\end{abstract}

\section{Introduction}
Sparse autoencoders offer promising insights for concept-based analysis of neural networks ~\cite{bricken2023towards, cunningham2023sparse}. Learning a sparse representation of a model's activations, they allow the extraction of interpretable features from both language and vision models.
Recent works compare different models upon SAE features, by constructing a common concept space ~\cite{thasarathan2025universal}, or by quantifying similarities between models ~\cite{wang2025towards}.
However, such works are restricted to a limited number of models (2 to 3), and only perform comparisons between models within the same modality.

We conduct a systematic analysis of 21 encoders with 3 datasets of text-image pairs, including both visual, textual and multimodal encoders, and compare them upon their inner concepts.
We introduce \textit{wMPPC} (\textit{weighted Max
Pairwise Pearson Correlation}), a similarity indicator between models, extending previous works ~\cite{wang2025towards} with emphasis on ``important'' features. Furthermore, we propose the \textit{Comparative Sharedness} of individual features, allowing the identification of features from a given model that are better shared with a class of model than another.

Our contributions thus include two new tools to interpret the inner representation of large neural networks (\autoref{sec:method}). With these tools, we also conduct a study involving 21 encoders that is not only much larger than previous works in this vein ~\cite{thasarathan2025universal,wang2025towards}, but also specifically deals with the multimodal aspect, comparing visual, textual and multimodal encoders, at different sizes (\autoref{sec:all_expes}). 
Originally, we also use several datasets as input, in particular to study the effect of a specific domain in the context of SAE-based interpretability. 
The main remarkable outcomes of this study include that 
(i) The shared information between models of different modalities is to be found mostly in the last layer of each model (\autoref{sec:wMPPC_model_level})
(ii) \textit{wMPPC} reveals differences in image-text alignment quality between datasets (\autoref{sec:alt_input}).
(iii) We establish a typology of SAE features learnt on CLIP's visual encoder that are shared with multiple VLMs, better than with classical visual foundation models (\autoref{subsec:visual-features-vlms}). Such features are related to high-level semantic concepts, such as specific geographical regions, or even purely textual information.
(iv) We find this typology to be similar to the one obtained while looking for visual features of CLIP that are better shared with text encoders (using image captions) than with visual foundation models (\autoref{sec:cs-llms}). Therefore, we highlight the impact of text pretraining on image understanding, by isolating individual concepts that are specific to those models.

\section{Method} \label{sec:method}
\subsection{SAEs}\label{sec:sae}
We train sparse autoencoders on models' activations, in order to learn specific features corresponding to interpretable semantic concepts. Each sparse autoencoder consists of two linear layers, and has the training objective of reconstructing its inputs. We use TopK sparse autoencoders \cite{gao2025scaling, makhzani2013k}, that directly constrain sparsity via an activation function, by only keeping the $k$ highest activations, and setting others to zero. This sparsity constraint makes the control of the sparsity level easier and more readable than other techniques using a $L_1$ penalization. Therefore, it enables training sparse autoencoders on multiple models' activations in a unified framework, even if the activations have different distributions.

With $W_{enc}, W_{dec} \in \mathbb{R}^{D \times F} \times \mathbb{R}^{F \times D}$ the respective weights of the SAE encoder and decoder, $b_{dec}$ its decoder bias and $x$ the studied model activations at a given layer, the SAE intermediate latent vector $\bm{f} \in \mathbb{R}^F$ is defined by:

\begin{equation}
    \bm{f} = W_{enc} \cdot (\bm{x} - \bm{b}_{dec})
    \label{eq:sae-features}
\end{equation}

This vector will further be referred as ``features'', with each feature representing a distinct semantic concept. These features are recorded and analyzed prior to applying the sparse TopK operator to ensure that no latent feature is overlooked.

\begin{equation}
    \hat{\bm{x}} = \textit{TopK}(\bm{f}) \cdot W_{dec} + \bm{b}_{dec}
    \label{eq:sae-reconstruction}
\end{equation}

Finally, the SAE reconstructs its input as $\hat{\bm{x}}$ with a MSE loss $\mathcal{L} =  ||\bm{x}-\hat{\bm{x}} ||^2$. 
It is trained on every token or patch (when applicable). However, at inference, we only consider the features corresponding to the global representation of a data sample (e.g. CLS token), in order to compare SAE features of models with different patch size, tokenizers, or even different modalities.

\subsection{Weighted MPPC}\label{sec:wMPPC}

In order to compare different models upon their SAE features (\autoref{eq:sae-features}), we extend the MPPC indicator \cite{wang2025towards}. The MPPC of the $i$-th SAE feature learnt for a pretrained model $A$ is defined by its maximum pairwise Pearson correlation among features of model $B$ and is noted $\rho_i^{A \rightarrow B}$:
\begin{equation}
    \rho_i^{A\rightarrow B} = \max_{j} \frac{\mathbb{E}[(\bm{f}_i^A - \mu_i^A)(\bm{f}_j^B - \mu_j^B)]}{\sigma_i^A \sigma_j^B}
    \label{rho-eq}
\end{equation}
with $\bm{f}_i^A, \bm{f}_j^B$ the $i$-th feature of $A$ and the $j$-th feature of $B$, $\mu_i^A, \mu_j^B$ their respective means, $\sigma_i^A, \sigma_j^B$ their standard deviations. In practice the correlations are estimated with $N$ sample data. At a model-scale, \cite{wang2025towards} proposes to define $\textit{MPPC}^{A \rightarrow B}$ as the arithmetic mean of $\rho_i^{A\rightarrow B}$ across all $m$ features of $A$ to assess the extent to which the features of $A$ are shared with $B$.

\begin{equation}
    \textit{MPPC}^{A \rightarrow B} = \frac{1}{m} \sum_{i=1}^m \rho_i^{A \rightarrow B}
\end{equation}

We hypothesize that some features are more important than others when quantifying similarities between two models. For instance, a feature $\bm{f}_i$ that is activated more frequently and with higher magnitude than another feature $\bm{f}_j$ may provide meaningful insight into a model’s behavior. We denote $S_i^A$ the cumulative activation of the $i$-th feature in model $A$ over a dataset $\mathcal{D}$.

\begin{equation}
    S_i^A=\sum_{\bm{x}\in \mathcal{D} }\bm{f}_i^A(\bm{x})
    \label{eq:wmppc-weight-s}
\end{equation}

Experimentally, the weight $S_i^A$ has a high variability. For example, for sparse autoencoders trained on the vision encoder of CLIP ViT-L/14, its coefficient of variation is $\frac{\sigma}{\mu} = 1.91$ . Also, considering \textit{MPPC} from CLIP to SigLIP2, we observe a significative correlation of 0.36 between $S_i$ and $\rho_i$ (\autoref{rho-eq}), suggesting that features with a high $S_i^A$ tend to be better shared than others. 

Using this measure of relative importance of SAE features, we introduce $\textit{wMPPC}$ (weighted \textit{MPPC}), that uses $S_i^A$ as a weighting factor.

\begin{equation}
    \textit{wMPPC}^{A \rightarrow B} = \frac{\sum_{i=1}^M S_i^A \cdot \rho_i^{A\rightarrow B} }{\sum_{i=1}^M S_i^A}
\end{equation}

Note that both $\textit{MPPC}$ and $\textit{wMPPC}$ are asymmetric and therefore do not constitute proper metrics. However, they effectively quantify the extent to which semantic concepts (i.e., features $\bm{f}_i$) of one model are shared with another. 

\subsection{Comparative Sharedness to identify individual features of a model}\label{sec:CS}

Using $\textit{wMPPC}$, we are able to evaluate whether two models share the same features on average, at a global scale. But a concept-level analysis can give even more insight on the inner representations of models, by establishing a typology of concepts that are specific to a group of models, but not shared with another. For a given feature $\bm{f}_i$ from a model $M$, we define the \textit{Comparative Sharedness} $\Delta_i^{M \rightarrow A,B}$ by:
\begin{equation}
    \Delta_i^{M \rightarrow A,B} = S_i^M  \times ({\rho_i^{M \rightarrow A}} - {\rho_i^{M \rightarrow B}}) ({\rho_i^{M \rightarrow A}} + {\rho_i^{M \rightarrow B}})
\end{equation}
where $S_i^M=\sum_{\bm{x}\in \mathcal{D} }\bm{f}_i(\bm{x})$ is computed over all the input images to weight the importance of the feature. Hence, $\Delta_i^{M \rightarrow A,B}$ is the difference of $\textit{wMPPC}$ contributions of the considered feature $\bm{f}_i$ of a model $M$ towards $A$ and $B$  (measuring if the feature is ``better shared`` with $A$ than with $B$), multiplied by the sum of their maximum correlations, in order to favour features that have high correlations with at least one model. This way, the features of $M$ with a high value of $\Delta_i^{M \rightarrow A,B}$ are ``well shared`` with $A$, but not with $B$.

In a cross-modal context, it is for instance interesting to identify the features of a visual encoder $M$ which are highly correlated to the textual features of a model $A$ but not to those of another visual encoder $B$. The approach is not restricted to the comparison to a couple of models and can be extended to two groups of models $\bm{G}$ and $\bm{H}$. To find features shared with every model in $\bm{G}$, but with no model in $\bm{H}$, we define the \textit{Generalized Comparative Sharedness}:
\begin{equation}
    \Delta_i^{M \rightarrow \bm{G},\bm{H}} = S_i^M \times \left((\min_{G_i \in \bm{G}} \rho_i^{M \rightarrow G_i})^{2} - (\max_{H_i \in \bm{H}} \rho_i^{M \rightarrow H_i})^{2}\right)
    \label{delta-groups-eq}
\end{equation}
In the vein of the example above, if we consider a visual encoder $M$, a set $\bm{T}$ of several textual encoders and a set $\bm{V}$ of several visual encoders,  high values of $\Delta_i^{M \rightarrow \bm{T},\bm{V}}$ would be associated with the features of $M$ that are specifically correlated to some textual features (among a large set) while being different from other visual features.

\subsection{On computational tractability}\label{sec:compute-tractability}
Computing $\textit{MPPC}^{A \rightarrow B}$, hence $\textit{wMPPC}^{A \rightarrow B}$, requires computing $n_A \times n_B$ Pearson correlations, with $n_A$ and $n_B$ the number of SAE features learnt on models $A$ and $B$. In practice, computing $\textit{wMPPC}^{A \rightarrow B}$ on every layer of models requires tens of billions of correlations, between vectors as long as the dataset.
The Pearson correlation $r(X,Y) = \frac{\mathbb{E}[(X-\mu_X)(Y-\mu_Y)]}{\sigma_X \sigma_Y}$ is computed from N samples of the random variables $X$ and $Y$. With $\tilde{X} = \frac{X - \mu_X}{\sigma_X}, \tilde{Y} = \frac{Y - \mu_Y}{\sigma_Y}$, we have $r(X,Y) = \frac{\tilde{X} \cdot \tilde{Y}}{N}$. Therefore, the Pearson correlation can be seen as a dot product between standardized vectors.
All the correlations required by \textit{wMPPC} can be computed in a single matrix multiplication, between the matrices of standardized features. The use of block matrix multiplication (or chunking) can be used, to solve potential memory issues. For example, computing \textit{wMPPC} on COCO between two models with 24 layers $\times$ 8192 features (largest models of this study, see \autoref{sec:appendix_list_encoders}) requires $9.14 \times 10^{15}$ FLOPs, and 469s on a single Nvidia A100 GPU, using FP32 precision at peak theoretical throughput. In practice, 5 runs of this settings took $608.6\pm 5.9$ seconds on our cluster machine. 
Considering only the last layer of each model (like required to compute \textit{Comparative Sharedness}) divides the number of operations by the number of layers of each model. Using the same two models, it would require $1.59 \times 10^{13}$ FLOPs, 0.81s at peak throughput on a single A100 GPU, and $1.01\pm 0.01$ seconds on 2880 timed runs.

\section{Experiments}\label{sec:all_expes}

\subsection{Experimental setup}\label{sec:exp-setup}

\paragraph*{Models} We consider several classes of models, with different architectures and various sizes. For VLMs, we use CLIP~\cite{radford2021learning}, DFN~\cite{fang2024data} and SigLIP2~\cite{tschannen2025siglip}, each having a visual and a textual encoder; for language models, we consider BERT~\cite{devlin-etal-2019-bert} and DeBERTa~\cite{he2021deberta}; for visual foundation models (FM), we use DinoV2~\cite{oquab2023dinov2} and ViT~\cite{dosovitskiy2020image}. We also consider MambaVision~\cite{hatamizadeh2024mambavision} as a visual FM,
but its architecture is different from a succession of transformer blocks, with blocks comprising both Mamba mixers and self-attention. Therefore, we consider it only at the last layer, as the choice of the network stages to consider as ``layers'' could cause drastic and arbitrary modifications towards \textit{wMPPC} at a model-level. All these models were tested in different sizes, using the \textit{base} and \textit{large} models. More details on these models can be found in appendix~\ref{sec:appendix_list_encoders}.

\paragraph*{Datasets} We consider two general domain datasets: COCO~\cite{lin2014microsoft}, in particular the \texttt{train2017} split with \numprint{118287} images and corresponding captions, and a subset of \numprint{61642} image-text pairs from Laion-2B\footnote{From the dataset \url{https://huggingface.co/datasets/MayIBorn/laion_2b_en_subset_70666}, we collected the images that were still available at the given urls, resulting in the \numprint{61642} image-text pairs.}~\cite{Schuhmann2022_laion}. We also consider a dataset in a specific domain, with images and captions: Oxford-102 Flowers\footnote{\url{https://huggingface.co/datasets/efekankavalci/flowers102-captions}}~\cite{nilsback2008automated}, with \numprint{8189} image-text pairs.

\paragraph*{Implementation details} The datasets are used as input of the encoders and we use the  activations of their layers as training data of the sparse autoencoders. SAEs are thus learnt in the residual stream after each transformer block, for every model. SAEs are trained with the Adam optimizer, using $\beta_1 = 0.9$ and $\beta_2 = 0.999$. The learning rate is set to $5\cdot 10^{-5}$ for all configurations. Also, we initialize $W_{enc}$ as $W_{dec}^T$ as per~\cite{gao2024scaling}, in order to prevent ``dead latents'' (never activated features). 
Our SAEs use a TopK architecture, with $k$ = 32, meaning that training is achieved by using 32 sparse codes to represent every input. This value was chosen as it was the smallest power of 2 obtaining no dead latents on COCO with CLIP-ViT-L/14.

Finally, we use an expansion factor of 8 (same as ~\cite{thasarathan2025universal}), meaning that the intermediate representation of SAEs is 8 times as large as their input dimension.
All the SAEs of this study were trained using the same SAE hyperparameters, in order to perform a systematic analysis of their learnt features.

\subsection{Comparison at the model level} \label{sec:wMPPC_model_level}


\begin{table}[tb]
\centering
\caption{$\textit{wMPPC}^{\textit{source}\to \textit{target}}$ (all layers) on COCO, for 6 large image and text encoders}
\label{tab:wmppc-coco-all}

\begin{tabular}{@{}cl|lll|lll@{}}
\toprule
\multicolumn{2}{c|}{\multirow{2}{*}{\diagbox{Source}{Target}}}  & \multicolumn{3}{c|}{Image} & \multicolumn{3}{c}{Text} \\
&   & CLIP (I)   & SigLIP2 (I) & DinoV2 & CLIP (T) & SigLIP2 (T) & BERT  \\ \midrule
\multirow{3}{*}{\rotatebox{90}{Image}}& CLIP (I)                     & 1          & 0.446      & 0.486  & 0.209   & 0.131      & 0.194 \\
& SigLIP2 (I)                   & 0.514      & 1          & 0.509  & 0.272   & 0.171      & 0.251 \\
& DinoV2                       & 0.556      & 0.518      & 1      & 0.250   & 0.153      & 0.233 \\
\midrule
\multirow{3}{*}{\rotatebox{90}{Text}}& CLIP(T)     & 0.253      & 0.275      & 0.246  & 1       & 0.351      & 0.428 \\
& SigLIP2 (T) & 0.045      & 0.050      & 0.043  & 0.256   & 1          & 0.578 \\
& BERT        & 0.182      & 0.194      & 0.177  & 0.346   & 0.287      & 1     \\ \bottomrule
\end{tabular}

\end{table}

We compute \textit{wMPPC} between the image (I) and text (T) encoder of CLIP and SigLIP2, Dino v2 and BERT, using the large (L-size) version of each and COCO as input dataset. When we consider all the layers of these 6 encoders, the results are reported in \autoref{tab:wmppc-coco-all}. 
At a model level, comparisons between encoders with the same modality obtain much higher $\textit{wMPPC}$ than cross modality comparisons, even when considering the two encoders of a same VLM. 
SigLIP2 reaches state-of-the-art performance on various vision-language tasks ~\cite{tschannen2025siglip}. Both \textit{wMPPC} between its two encoders (0.050 and 0.171) are nevertheless lower than \textit{wMPPC} between BERT and DinoV2 (0.177 and 0.233).

\begin{figure}[htbp]
    \centering
    \includegraphics[width=0.75\textwidth]{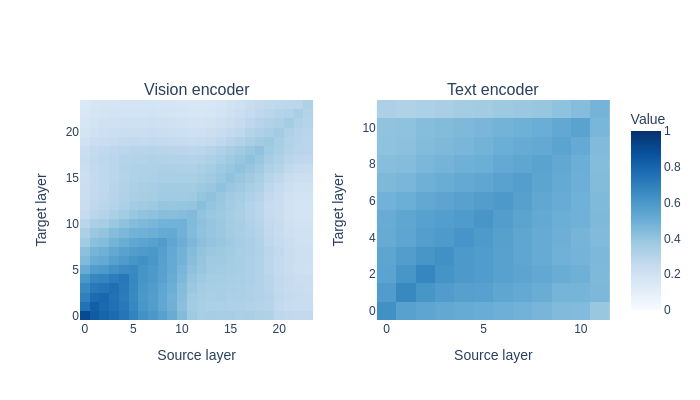}
    \caption{Layerwise \textit{wMPPC} between 2 SAEs trained on each encoder of CLIP}
    \label{fig:heatmaps-img-txt}
\end{figure}

However, nothing guarantees that the early layers of image and text encoders would correspond to features of the same semantic level. Therefore, we train two SAEs on each encoder of a CLIP-ViT-L/14 model. We then represent the layerwise $\textit{wMPPC}$ of the two encoders in \autoref{fig:heatmaps-img-txt}. 
The SAEs learnt on the text encoder obtain similar $\textit{wMPPC}$ for every pair of layers considered. For the image encoder, we encounter much higher $\textit{wMPPC}$ on early layers, and $\textit{wMPPC}$ stays concentrated along the diagonal, with layers of different levels obtaining low $\textit{wMPPC}$, hence representing very different features. Therefore, as suggested by previous works ~\cite{NEURIPS2023_abf3682c}, features with the highest semantic level should be found in the last layer of vision encoders.

\begin{table}[tb]
\centering
\caption{$\textit{wMPPC}^{\textit{source}\to \textit{target}}$ (last layers) on COCO, for 6 large image and text encoders}
\label{tab:wmppc-coco-last}

\begin{tabular}{@{}cl|lll|lll@{}}
\toprule
\multicolumn{2}{c|}{\multirow{2}{*}{\diagbox{Source}{Target}}}  & \multicolumn{3}{c|}{Image} & \multicolumn{3}{c}{Text} \\
&  & CLIP (I)   & SigLIP2 (I)     & DinoV2         & CLIP(T)        & SigLIP2 (T)     & BERT           \\ \midrule
\multirow{3}{*}{\rotatebox{90}{Image}}& CLIP (I) & 1 & 0.278 & 0.208 & 0.220 & 0.128 & 0.203 \\
& SigLIP2 (I) & 0.320 & 1  & 0.236  & 0.274  & 0.153  & 0.249   \\
& DinoV2      & 0.270 & 0.290  & 1     & 0.254 & 0.142 & 0.216   \\ \midrule
\multirow{3}{*}{\rotatebox{90}{Text}}& CLIP(T) & 0.255 & 0.284  & 0.211  & 1 & 0.192 & 0.286          \\
& SigLIP2 (T) & 0.054 & 0.062 & 0.042 & 0.134 & 1 & 0.297  \\
& BERT        & 0.183 & 0.195 & 0.136 & 0.237 & 0.172 & 1  \\ \bottomrule
\end{tabular}

\end{table}

We compute \textit{wMPPC} considering only the last layer of each model, with the results being reported in~\autoref{tab:wmppc-coco-last}. In that case, \textit{wMPPC} decreases substantially for same-modality comparisons (even for two text encoders), but stays stable or even increases for cross-modal comparisons. Therefore, we deduce that the shared information between models of different modalities is to be found mostly in the last layer of each model.

\begin{table}[htbp]
\caption{Average \textit{wMPPC} for all model pairs, combined by the modality of the source and target encoders, and tested on different datasets and for different model sizes. Each number correspond to the average score of a quarter of table such as \autoref{tab:wmppc-coco-all} (all) or \autoref{tab:wmppc-coco-last} (last), without the '1' on the diagonal and all models of each type (instead of 3 only in \autoref{tab:wmppc-coco-all} and \autoref{tab:wmppc-coco-last}), either in their large (L with 10 encoders) or basic (B with 10 other encoders) size. Detailed results for each model are provided in the Appendix.}
\label{tab:wmppc-different-setups}
\centering
\begin{tabular}{@{}lll|cccc@{}}
\toprule
 \multirow{2}{*}{Layers} & Input   & Models & & & & \\
 & dataset & size & Image $\rightarrow$ Image & Image $\rightarrow$ Text & Text $\rightarrow$ Image & Text $\rightarrow$ Text \\ 
\midrule
\multirow{4}{*}{All}  
& COCO & L      & 0.463 & 0.204 & 0.168 & 0.367 \\
& COCO & B      & 0.509 & 0.225 & 0.178 & 0.405 \\
& Laion & L     & 0.470 & 0.146 & 0.140 & 0.524 \\
& Flowers & L   & 0.548 & 0.180 & 0.129 & 0.447 \\
\midrule
\multirow{4}{*}{Last}
& COCO & L      & 0.265 & 0.213 & 0.173 & 0.249 \\
& COCO & B      & 0.281 & 0.222 & 0.176 & 0.275 \\
& Laion & L     & 0.187 & 0.119 & 0.122 & 0.308 \\
& Flowers & L   & 0.377 & 0.133 & 0.166 & 0.281 \\
\bottomrule
\end{tabular}

\end{table}

Furthermore, we conduct a similar analysis with four more encoders, namely the image and text encoder of DFN, a ViT (image encoder) and DeBERTa (text encoder), resulting in a total of 10 large image and text encoders. In~\autoref{tab:wmppc-different-setups}, we report the average of $\textit{wMPPC}^{A \rightarrow B}$ with  each encoder modality for $A$ and $B$. ``Image $\rightarrow$ Text'' represents the average \textit{wMPPC} with any image encoder as the source and a text encoder as the target (respectively for other combinations). Same-encoder comparisons are omitted from the average.

Using smaller versions of the same encoders (B-size instead of L-size) on COCO, \textit{wMPPC} appears to be very similar (or slightly higher) as shown in~\autoref{tab:wmppc-different-setups}, thus leading to the same conclusions as using L-size models. 

\subsection{Alternative input dataset}\label{sec:alt_input}
Previous studies dealing with SAE-based interpretability relied on a single input dataset to generate the activations on which the SAE are learnt. We propose using two other datasets to refine the previous analysis. The results are reported in~\autoref{tab:wmppc-different-setups} with 10 large models.

In order to analyze whether our previous observations transpose to another dataset, we compute \textit{wMPPC} on SAE features learnt on a subset of 61642 image-text pairs from Laion-2B ~\cite{Schuhmann2022_laion}. At a model level, \textit{wMPPC} values are similar to those obtained on COCO, except for comparisons between two text encoders, that obtain higher scores. However, cross modal comparisons obtain much lower \textit{wMPPC} when considering only the last layer. As the Laion-2B captions are scraped from the internet (as opposed to COCO's that are human-written for each image), this could highlight a worse image-text alignment for this dataset.

To compare models on a domain specific dataset, we compute \textit{wMPPC} between L-size models on Oxford-102 Flowers. As this dataset has less intra-modality variability (domain specific), \textit{wMPPC} gets higher scores than on COCO for same-modality comparisons, especially between image encoders. However, cross-modal comparisons obtain lower \textit{wMPPC} than on COCO, suggesting a worse image-text alignment.

\subsection{A typology of visual concepts specific to VLMs}
\label{subsec:visual-features-vlms}

The use of image-text contrastive learning has shown great results in understanding visual information. Then, we aim at exhibiting the gain made possible by such multimodal training, at a $\textit{concept}$ level, by using our SAE-based indicators. SAEs are trained on the activations resulting from the COCO dataset, holding a high image-text alignment quality (\autoref{sec:alt_input}). 
In order to identify features shared by multiple VLMs, but not by visual FMs,
we compute the Generalized Comparative Sharedness $\Delta^{M \rightarrow G,H}$  (\autoref{delta-groups-eq}), with CLIP features as a comparison standard $M$. For this role, we consider CLIP among all VLMs used in this study, as it is the most common, the oldest, and the least performant one. Therefore, features from CLIP that have low $\rho_i^{A \rightarrow B}$ towards other VLMs would not be explained by a performance improvement. The group $\bm{G}$ comprises the visual encoders from other VLMs (SigLIP2 and DFN) as well as  features from a second SAE trained on the same CLIP model as $M$, with a different seed. The group $\bm{H}$ comprises the visual foundation models DinoV2, MambaVision and a ViT trained on ImageNet-21k classification.

Inspecting the features corresponding to the top 1\% of $\Delta^{M \rightarrow G,H}$ (81 out of 8192), we establish the following typology of concepts that are specific to VLMs :

\begin{itemize}
    \item Age related features. Among the features that are specific to all the studied VLMs, some are associated to kids in specific situations, such as birthday parties, brushing teeth or playing baseball. Each of those features is associated with a specific age range.
    \item Pets having ``unusual'' behaviour. The COCO dataset has lots of images of cats and dogs having ``unusual'' behaviour, such as wearing ties or hats, sitting on laptops... VLMs share multiple features associated specifically to those images, often to multiple types of those ``unusual behaviours'', but not to classical images of pets. Visual FM don't share those features.
    \item Rooms of the house : features activated by images of a specific room of the house (bedroom, bathroom, kitchen...). In particular, some features with a high comparative sharedness. 
    are activated on images of different types of the bathroom (sink, toilet, bath). Also, those images are more cluttered than most of the COCO dataset, however coherent associations are made.
    \item Vehicles. High speed trains, fret trains and steam trains are all visually different, however they all are trains. Then, CLIP has features activated for all those kinds of trains, and similar features for planes, cars, buses or boats are shared with all the studied VLM, but with none of the studied visual FM.
    \item ``Old'' photos : features activated for grayscale, blurry, and seemingly old photos. Even though those characteristics are purely visual, those features are specific to VLMs. Also note that recent artistic grayscale photos are present in the dataset, and have distinct features associated to them, not obtaining a high comparative sharedness.
    \item Geographical features : features activated on different kinds of images corresponding to the same geographical region. That includes features activated for multiple types of african animals (elephants, zebras and giraffes), or multiple types of Italian food (such as pastas, lasagnas and pizzas).
    Note that features activated only for images of zebras, or only for pizzas do not obtain a high comparative sharedness.
    \item ``To ride'' : one notable feature among the top 1\% of $\Delta^{M \rightarrow G,H}$ is activated for images of horses, skis, snowboards, bikes, surfs or jetskis. Those are very different types of objects, but all of them are associated to the verb "to ride" in English. 
\end{itemize}

Such observations confirm previous assumptions on geographical features ~\cite{stevens2025sparse}, but allows extracting a whole typology of concepts by having a more systematic approach.
Most of those features seem to rely on prior knowledge, that is absent from visual foundation models without text pretraining. They are activated on images of different types of situations, corresponding to the same high-level semantic concept.
In particular, the feature seemingly related to the verb ``to ride'' appears to rely solely on textual information, despite being extracted from a visual encoder.

\subsection{Visual features specific to VLMs are really \textit{textual} features}
\label{sec:cs-llms}

We established a typology of SAE features that are shared by multiple VLMs, but not by visual FMs. Then, if these specificities are a direct consequence of text pre-training, some features learnt on text encoders using image captions could have similar behaviours. We study the same CLIP image features as previously, using Generalized Comparative Sharedness $\Delta^{M \rightarrow G,H}$  to find features better shared with BERT-large and DeBERTa-large than with any of MambaVision, DinoV2 and ViT. Again, we establish a typology of concepts among the top 1\% of $\Delta^{M \rightarrow G,H}$ . The features of CLIP's image encoder that are better shared with every studied LLM than with any studied visual FM correspond to: ``kids in a specific situation'', ``rooms of the house'', ``types of vehicles'', ``pets having unusual behaviour'' or ``old photos''. 

\begin{figure}[ht]
  \centering
  \begin{subfigure}{0.45\textwidth}
    \centering
    \includegraphics[width=\linewidth]{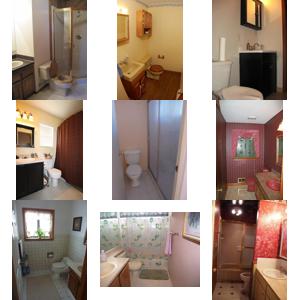}
  \end{subfigure}
  \hspace{0.05\textwidth}
  \begin{subfigure}{0.45\textwidth}
    \centering
    \includegraphics[width=\linewidth]{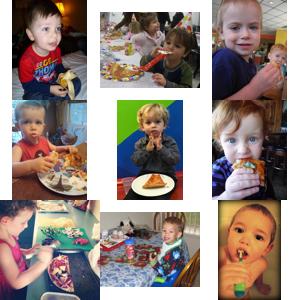}
  \end{subfigure}
  \caption{CLIP visual features better shared with LLMs and VLMs than with visual FMs}
  \label{fig:cs-textual-examples}
\end{figure}

The obtained typology is very similar to the one established while considering VLMs' visual encoders, pushing the hypothesis that previous observations could be caused by their text pretraining. Actually, 16 features are present in the highest 81 (1\%) {Comparative Sharedness} towards both LLMs and VLMs' visual encoders. Qualitative examples are represented in \autoref{fig:cs-textual-examples}, with the 9 images corresponding to the highest activations among the COCO dataset.

\section{Related work}

\paragraph{Representational similarity}
As both the performance of deep neural networks improves on both text and images, recent works analyze the alignment between the representations of such networks ~\cite{kornblith2019similarity, klabunde2023similarity, boix2022gulp}. Empirical studies ~\cite{li2024do_vlm_share_concepts, huh2024platonic} find representational alignment between language and vision models, by studying the distance structure induced by their learnt vector embeddings. In particular, they find convergence of models of different architectures and modalities upon performance, suggesting the existence of a \textit{platonic representation}.

\paragraph{Universal neurons}
Analyzing individual neurons of networks has revealed neurons corresponding to interpretable features. In language models, some have been found to correspond to sentiment ~\cite{donnelly2019interpretability} or skills ~\cite{wang2022finding}. In a similar fashion, vision models have individual neurons activated for curves with specific orientations ~\cite{cammarata2021curve} or specific objects ~\cite{bau2020understanding}.
Multiple GPT2 models trained with different training seeds have been shown to share 1-5\% of neurons ~\cite{gurnee2024universal}, with clear interpretations and functional roles. Also, vision models trained with different tasks share \textit{Rosetta Neurons}, activated on similar regions of images ~\cite{dravid2023rosetta}.
Semantic superposition is the main problem for such studies, as most neurons are polysemantic, and are activated on seemingly unrelated inputs ~\cite{elhage2022toy}.

\paragraph{Sparse autoencoders}

In order to disentangle the concepts corresponding to individual neurons, sparse autoencoders are trained on models' activations, in order to extract sparse, and interpretable features ~\cite{bricken2023towards, cunningham2023sparse}.
Such features have seen promising results towards understanding language models ~\cite{lieberum2024gemma, gao2025scaling, rajamanoharan2024improving}.
Also, recent works have addressed SAEs for vision, or multimodal models ~\cite{gorton2024missing, thasarathan2025universal, lim2025sparse, rao2024discover}, in scenarios such as model adaptation ~\cite{lim2025sparse}.
~\cite{NEURIPS2023_abf3682c} evaluates the importance of each learnt concept, by assessing its impact on classification predictions.

\paragraph{Comparing SAEs}
SAE features are used to compare different models. Universal SAEs ~\cite{thasarathan2025universal} learn a common concept space for three image encoders, relying on the same decoder. \textit{MPPC} ~\cite{wang2025towards} performs a correlation analysis between features of two generative LLMs, in order to quantify to what extent those models share concepts. Finally, ~\cite{stevens2025sparse} suggests that CLIP holds visual features associated with a precise cultural or geographical context, that are absent from DinoV2.

\section{Discussion, limitations and perspectives} \label{sec:limitations}

\paragraph{Discussion}
We conduct a comparative analysis of 21 visual, textual and multimodal encoders upon SAE-derived features. We introduce \textit{wMPPC}, an indicator evaluating similarities between different models at a \textit{concept} level, considering relative feature importance. Using it, we find that SAEs learnt on COCO obtain higher \textit{wMPPC} between encoders of different modalities than SAEs learnt on a subset of Laion-2B. That highlights the difference in quality of image-text alignment between the two datasets. Also, our \textit{Comparative Sharedness} indicator allows us to find individual features of a model that are better shared with a class of models than another one, and to establish typologies of such features. We find that features that are specific to VLMs among vision encoders are also better shared with LLMs than visual foundation models. That emphasizes the importance of text pretraining for image understanding, by highlighting specific concepts.

\paragraph{Limitations}
Although our study involves much more encoders than previous studies, all of them are based on transformers~\cite{vaswani2017transformers}. 
Training SAEs on models having large and hierarchical feature maps (such as convolutional networks, or Swin transformers ~\cite{liu2021swin}) is possible. However, in practice such models would imply having huge SAEs, or using smaller SAEs for the largest layers ~\cite{gorton2024missing}, therefore not allowing a systematic \textit{wMPPC} analysis. One can also note that we considered only encoders while MPPC~\cite{wang2025towards} focuses on language decoders. Our choice resulted from the initial aim of studying the features shared in a multimodal context. Although visual generative models conditioned by a text could have been considered, it seemed more appropriate to first study Visual Language Models which are trained with an objective that is more symmetric between both modalities. A second limitation is the asymmetry of the proposed \textit{wMPPC}, in the vein of the previous MPPC. Hence, it can not be used as a direct distance measurement between models. A naive symmetric version can be easily derived (\textit{e.g} similarly to the Jensen–Shannon divergence with regard to the Kullback-Leibler one) but it would be at the cost of losing important information. For instance in a cross-modal context, the wMPPC is very low when SigLIP is used as source but 3 times larger when it is used as target (\autoref{tab:wmppc-coco-all}). It suggests that SigLIP encodes more concepts that are unknown by image encoders than the opposite. A symmetric version of wMPPC would not be able to highlight such a phenomenon, reporting only a bland average value of both cross-modal contexts. A final limitation identified is that even if sparse autoencoders are one of the most promising methods regarding concept-based analysis, they are not guaranteed to extract every single concept used by a model. 

\paragraph{Broader impact}
By addressing specifically XAI in a cross-modal context, this paper can contribute to transfer representation from one modality to another. It can also contribute to improving a user's understanding of the inner structure of a large model, by providing explanations through multiple modalities. 

\paragraph{Perspectives}
Our findings highlight that \textit{wMPPC} can be used to assess the quality of the image-text alignment of a dataset. Comparative studies of multiple image-text datasets could be performed, in order to select or filter datasets used for training new models. 
Techniques for automatically naming SAE features considering both images and captions could allow large scale \textit{Comparative Sharedness} analysis, using features of both modalities as comparison standards.
All the models considered in this work are encoder models. As SAE-derived features have been studied extensively for models specialized in text generation, a systematic analysis of \textit{wMPPC} on generative models with different modalities could provide meaningful insight into their behaviour.

\section*{Acknowledgments}
This work was partially funded by the Agence Nationale de la Recherche (ANR) for the STUDIES project  ANR-23-CE38-0014-02.

\medskip

{
\small
\bibliography{ref}
}

\newpage

\appendix

\section{Appendix: encoder description}\label{sec:appendix_list_encoders}
We report in \autoref{tab:encoder_description} all the encoder considered in our study with their key features. All the models were downloaded from huggingface, except from CLIP and DFN models from OpenClip ~\cite{ilharco_gabriel_2021_5143773} and DinoV2 from PyTorch Hub. The \textit{model size} is the number of parameters and since all of them were encoded in \texttt{float32} their actual size in memory is this number multiplied by four. 

CLIP was trained ``on publicly available image-caption data'' that is images-caption pairs from the Web and publicly available datasets such as YFCC 100M~\cite{thomee2016yfcc}. The creator of the model did not release the dataset to avoid its use ``as the basis for any commercial or deployed model''.

DFN is a CLIP-like model trained from 2B image-text pairs, resulting from the filtering of a pool of 12.8 billion uncurated image-text pairs of CommonPool, collected from Common Crawl. This last is itself part of DataComp, a benchmark for designing multimodal datasets~\cite{gadre2023datacomp}.

MambaVision and ViT were trained on well-known and publicly available ImageNet dataset ~\cite{deng2009imagenet} with $1,000$ categories of the Large Scale Visual Recognition Challenge~\cite{russakovsky2015imagenet} or the full $21k$ classes.

In \autoref{tab:wmppc-different-setups}:
\begin{itemize}
    \item the set ``L`` contains 10 encoders: CLIP ViT L/14 (both image and text encoders), DFN ViT L/14 (both image and text encoders), SigLIP2 L/16 (both image and text encoders), DinoV2 L/14 (image encoder), ViT L/16 (image encoder), BERT large (text encoder) and DeBERTa (text encoder)
    \item the set ``B`` contains 10 encoders: CLIP ViT B/32 (both image and text encoders), DFN ViT B/16 (both image and text encoder), SigLIP2 B/16 (both image and text encoder), DinoV2 B/14 (image encoder), ViT B/16 (image encoder), BERT base (text encoder) and DeBERTa base (text encoder)
\end{itemize}

\begin{table}[hb]
    \centering
    \caption{Pre-trained encoders considered in this study.}
    \label{tab:encoder_description}
    \begin{tabular}{llllp{5cm}}
        \toprule
        Encoder & input  & Model &Model & Training set \\
         & type & type & size & \\
        \midrule
         CLIP ViT B/32 ~\cite{radford2021learning} \href{https://huggingface.co/openai/clip-vit-base-patch32}{\faDownload} & image & VLM & 87M &  \tikzmark{top1}\multirow{2}{*}{\quad openAI private: web, YFCC100M...}\\
         CLIP ViT L/14 ~\cite{radford2021learning} \href{https://huggingface.co/openai/clip-vit-large-patch14}{\faDownload}& image & VLM & 303M & \tikzmark{bot1}\\
         DFN ViT B/16 ~\cite{fang2024data} \href{https://huggingface.co/apple/DFN2B-CLIP-ViT-B-16}{\faDownload}& image & VLM & 86M & \tikzmark{top2}\multirow{2}{*}{\quad \parbox{5cm}{2B filtered from CommonPool-12.8B}} \\
         DFN ViT L/14 ~\cite{fang2024data} \href{https://huggingface.co/apple/DFN2B-CLIP-ViT-L-14}{\faDownload}& image & VLM & 303M & \tikzmark{bot2} \\
         SigLIP2 B/16 ~\cite{tschannen2025siglip} \href{https://huggingface.co/google/siglip2-base-patch16-224}{\faDownload}& image & VLM & 92M & \tikzmark{top3}\multirow{2}{*}{\quad WebLI} \\
         SigLIP2 L/16 ~\cite{tschannen2025siglip} \href{https://huggingface.co/google/siglip2-large-patch16-384}{\faDownload} & image & VLM & 316M & \tikzmark{bot3} \\
         DinoV2 B/14 ~\cite{oquab2023dinov2} \href{https://dl.fbaipublicfiles.com/dinov2/dinov2_vitb14/dinov2_vitb14_pretrain.pth}{\faDownload} & image & visual FM& 86M & \tikzmark{top4} \multirow{2}{*}{\quad LVD-142M} \\
         DinoV2 L/14 ~\cite{oquab2023dinov2} \href{https://dl.fbaipublicfiles.com/dinov2/dinov2_vitb14/dinov2_vitl14_pretrain.pth}{\faDownload} & image & visual FM& 304M & \tikzmark{bot4}  \\
         MambaVision B ~\cite{hatamizadeh2024mambavision} \href{https://huggingface.co/nvidia/MambaVision-B-1K}{\faDownload} & image & visual FM & 97M & \tikzmark{top5}\multirow{2}{*}{\quad ImageNet-1k} \\
         MambaVision L ~\cite{hatamizadeh2024mambavision} \href{https://huggingface.co/nvidia/MambaVision-L-1K}{\faDownload} & image & visual FM & 227M & \tikzmark{bot5} \\
         ViT B/16 \href{https://huggingface.co/google/vit-base-patch16-384}{\faDownload} ~\cite{dosovitskiy2020image} \href{https://huggingface.co/google/vit-base-patch16-224}{\faDownload} & image & visual FM & 86M &  \tikzmark{top6} \multirow{2}{*}{\quad ImageNet-21k} \\
         ViT L/16 ~\cite{dosovitskiy2020image} \href{https://huggingface.co/google/vit-large-patch16-224}{\faDownload} & image & visual FM & 303M  & \tikzmark{bot6}  \\
         BERT base  ~\cite{devlin-etal-2019-bert} \href{https://huggingface.co/google-bert/bert-base-uncased}{\faDownload}& text & LLM & 109M &  \tikzmark{top7} \multirow{2}{*}{\quad BookCorpus, Wikipedia} \\
         BERT large ~\cite{devlin-etal-2019-bert} \href{https://huggingface.co/google-bert/bert-large-uncased}{\faDownload}& text & LLM & 335M &  \tikzmark{bot7} \\
         DeBERTa base ~\cite{he2021deberta} \href{https://huggingface.co/microsoft/deberta-base}{\faDownload} & text & LLM & 99M & \tikzmark{top8}\multirow{2}{*}{\quad \parbox{5cm}{BookCorpus, Wikipedia, OpenWebText, STORIES}} \\
         DeBERTa large ~\cite{he2021deberta} \href{https://huggingface.co/microsoft/deberta-large}{\faDownload} & text & LLM & 353M &  \tikzmark{bot8}  \\
        \bottomrule
    \end{tabular}
    \DrawVerticalBrace[black, thick]{top1}{bot1}%
    \DrawVerticalBrace[black, thick]{top2}{bot2}%
    \DrawVerticalBrace[black, thick]{top3}{bot3}%
    \DrawVerticalBrace[black, thick]{top4}{bot4}%
    \DrawVerticalBrace[black, thick]{top5}{bot5}%
    \DrawVerticalBrace[black, thick]{top6}{bot6}%
    \DrawVerticalBrace[black, thick]{top7}{bot7}%
    \DrawVerticalBrace[black, thick]{top8}{bot8}%
\end{table}

\clearpage
\section{Appendix: statistical significance of MPPC and wMPPC}\label{sec:stat_signif}

With $\rho_i$ the maximum pairwise coefficient (\autoref{rho-eq}) for $N$ target features of length $L$, and $H_0$ the hypothesis of features having no linear relationship. Using the Fischer z-transformation \cite{fisher1915frequency}

$$
z = artanh(r) \sim \mathcal{N}(0, \frac{1}{\sqrt{L - 3}})
$$
$$
\mathbb{P}(\max_{i}(r_i) > x) = 1 - \mathbb{P}( r \le x)^N
$$
$$
\mathbb{P}(\rho_i > x) = \mathbb{P}(\max_{i}(z_i) > artanh(x)) 
$$
$$
\mathbb{P}(\rho_i > x) = 1 - \Phi(artanh(x) \sqrt{L-3}) ^N
$$

With $N = 8192$ (corresponding to ViT-L experiments, for one layer), and $L=10000$ being largely lower than the size of the COCO dataset used, we obtain $\mathbb{P}(\rho_i > 0.3) \approx 10^{-206}$ , thus reject $H_0$.

Experimentally, for two sparse autoencoders trained on the same CLIP-ViT-B/32 visual encoder, on the COCO dataset and shuffling the features upon images (to preserve the density of feature distributions), we obtain $\textit{wMPPC} = 0.0125$, while the non-shuffled $\textit{wMPPC} = 0.5854$.

\clearpage
\section{Appendix: detailed wMPPC results}
In \autoref{tab:wmppc-different-setups} we report average results of wMPPC that provide an overview to the reader for various settings. Below, we report the detailed results of each setting that led to these average scores. The setting are ``multidimensional'' thus we provide in \autoref{tab:pointer_to_detailed_results} the synthetic pointers to help the reading:

\begin{table}[hb]
    \centering
    \caption{Synthetic pointers to the tables of detailed results. It gives the table according to the input dataset and the size of the encoders (each row), the modality of the target enccoders that can be `Image' (col 3) or `Text' (col 4) and finally whether `all' layers or only the `last' one is used to compute wMPPC.}
    \label{tab:pointer_to_detailed_results}
    \begin{tabular}{ll|cc}
    \toprule
    \multirow{2}{*}{Dataset}& \multirow{2}{*}{Model size}  & \multicolumn{2}{c}{Target encoders} \\
     &  & Image & Text  \\
    \midrule
    \multirow{2}{*}{COCO} & large & \autoref{tab:wmppc-coco-all-10-img} (all) \autoref{tab:wmppc-coco-last-10-img} (last) &  \autoref{tab:wmppc-coco-all-10-txt} (all) \autoref{tab:wmppc-coco-last-10-txt} (last)\\
     & base &  \autoref{tab:wmppc-coco-base-all-10-img} (all) \autoref{tab:wmppc-coco-base-last-10-img} (last) &  \autoref{tab:wmppc-coco-base-all-10-txt} (all) \autoref{tab:wmppc-coco-base-last-10-txt} (last) \\
    LAION & large & \autoref{tab:wmppc-laion-large-all-10-img} (all) \autoref{tab:wmppc-laion-large-last-10-img} (last) &  \autoref{tab:wmppc-laion-large-all-10-txt} (all) \autoref{tab:wmppc-laion-large-last-10-txt} (last)\\
    Flowers-102 & large & \autoref{tab:wmppc-flowers-large-all-10-img} (all) \autoref{tab:wmppc-flowers-large-last-10-img} (last) &  \autoref{tab:wmppc-flowers-large-all-10-txt} (all) \autoref{tab:wmppc-flowers-large-last-10-txt} (last)\\
    \bottomrule
    \end{tabular}
\end{table}

Also, note that MambaVision is only considered at its last layer on COCO (\autoref{tab:wmppc-coco-last-10-img} and \autoref{tab:wmppc-coco-last-10-txt}), as it is only used to compute Comparative Sharedness.

\begin{table}[hb]
\centering
\caption{$\textit{wMPPC}^{\textit{source}\to \textit{target}}$ (all layers) on COCO, for all 10 large models as source, image encoders as target}
\label{tab:wmppc-coco-all-10-img}
\begin{tabular}{@{}cl|ccccc@{}}
\toprule
\multicolumn{2}{c|}{\multirow{2}{*}{\diagbox[dir=NW]{Source}{Target}}}
    & \multicolumn{5}{c}{Image} \\
&   & CLIP (I) & SigLIP2 (I) & DFN (I) & DinoV2 & ViT \\
\midrule
\multirow{5}{*}{\rotatebox{90}{Image}}
& CLIP (I)    & 1     & 0.446 & 0.489 & 0.486 & 0.444 \\
& SigLIP2 (I) & 0.514 & 1     & 0.516 & 0.509 & 0.500 \\
& DFN (I)     & 0.469 & 0.416 & 1     & 0.431 & 0.385 \\
& DinoV2      & 0.556 & 0.518 & 0.533 & 1     & 0.515 \\
& ViT         & 0.390 & 0.381 & 0.379 & 0.391 & 1     \\
\midrule
\multirow{5}{*}{\rotatebox{90}{Text}}
& CLIP (T)    & 0.253 & 0.275 & 0.254 & 0.246 & 0.223 \\
& SigLIP2 (T) & 0.045 & 0.051 & 0.044 & 0.043 & 0.037 \\
& DFN (T)     & 0.248 & 0.282 & 0.257 & 0.235 & 0.227 \\
& BERT        & 0.182 & 0.195 & 0.181 & 0.177 & 0.158 \\
& DeBERTa     & 0.119 & 0.129 & 0.120 & 0.113 & 0.105 \\
\bottomrule
\end{tabular}
\end{table}

\begin{table}[hb]
\centering
\caption{$\textit{wMPPC}^{\textit{source}\to \textit{target}}$ (all layers) on COCO, for all 10 large models as source, text encoders as target}
\label{tab:wmppc-coco-all-10-txt}
\begin{tabular}{@{}cl|ccccc@{}}
\toprule
\multicolumn{2}{c|}{\multirow{2}{*}{\diagbox[dir=NW]{Source}{Target}}}
    & \multicolumn{5}{c}{Text} \\
&   & CLIP (T) & SigLIP2 (T) & DFN (T) & BERT & DeBERTa \\
\midrule
\multirow{5}{*}{\rotatebox{90}{Image}}
& CLIP (I)    & 0.209 & 0.131 & 0.214 & 0.194 & 0.188 \\
& SigLIP2 (I) & 0.272 & 0.171 & 0.282 & 0.251 & 0.248 \\
& DFN (I)     & 0.203 & 0.128 & 0.208 & 0.188 & 0.182 \\
& DinoV2      & 0.250 & 0.153 & 0.256 & 0.233 & 0.224 \\
& ViT         & 0.201 & 0.132 & 0.207 & 0.186 & 0.183 \\
\midrule
\multirow{5}{*}{\rotatebox{90}{Text}}
& CLIP (T)    & 1     & 0.351 & 0.509 & 0.428 & 0.412 \\
& SigLIP2 (T) & 0.256 & 1     & 0.254 & 0.578 & 0.400 \\
& DFN (T)     & 0.480 & 0.327 & 1     & 0.426 & 0.431 \\
& BERT        & 0.346 & 0.287 & 0.352 & 1     & 0.361 \\
& DeBERTa     & 0.266 & 0.256 & 0.274 & 0.344 & 1     \\
\bottomrule
\end{tabular}
\end{table}

\begin{table}[tb]
\centering
\caption{$\textit{wMPPC}^{\textit{source}\to \textit{target}}$ (last layer) on COCO, for all 10 large models and MambaVision as source, image encoders as target}
\label{tab:wmppc-coco-last-10-img}
\begin{tabular}{@{}cl|cccccc@{}}
\toprule
\multicolumn{2}{c|}{\multirow{2}{*}{\diagbox[dir=NW]{Source}{Target}}}
    & \multicolumn{6}{c}{Image} \\
&   & CLIP (I) & SigLIP2 (I) & DFN (I) & DinoV2 & ViT & MambaVision\\
\midrule
\multirow{5}{*}{\rotatebox{90}{Image}}
& CLIP (I)    & 1     & 0.278 & 0.265 & 0.208 & 0.232 & 0.215\\
& SigLIP2 (I) & 0.320 & 1     & 0.329 & 0.236 & 0.293 & 0.294 \\
& DFN (I)     & 0.267 & 0.287 & 1     & 0.207 & 0.236 & 0.225 \\
& DinoV2      & 0.270 & 0.290 & 0.277 & 1     & 0.270 & 0.259 \\
& ViT         & 0.258 & 0.286 & 0.272 & 0.226 & 1     & 0.264 \\
& MambaVision & 0.236 & 0.281 & 0.258 & 0.214 & 0.264 & 1     \\
\midrule
\multirow{5}{*}{\rotatebox{90}{Text}}
& CLIP (T)    & 0.255 & 0.284 & 0.265 & 0.211 & 0.252 & 0.234 \\
& SigLIP2 (T) & 0.054 & 0.062 & 0.054 & 0.042 & 0.049 & 0.048 \\
& DFN (T)     & 0.246 & 0.287 & 0.258 & 0.196 & 0.245 & 0.227 \\
& BERT        & 0.183 & 0.195 & 0.179 & 0.136 & 0.168 & 0.154 \\
& DeBERTa     & 0.150 & 0.162 & 0.149 & 0.115 & 0.139 & 0.128 \\
\bottomrule
\end{tabular}
\end{table}

\begin{table}[tb]
\centering
\caption{$\textit{wMPPC}^{\textit{source}\to \textit{target}}$ (last layer) on COCO, for all 10 large models as source, text encoders as target}
\label{tab:wmppc-coco-last-10-txt}
\begin{tabular}{@{}cl|ccccc@{}}
\toprule
\multicolumn{2}{c|}{\multirow{2}{*}{\diagbox[dir=NW]{Source}{Target}}}
    & \multicolumn{5}{c}{Text} \\
&   & CLIP (T) & SigLIP2 (T) & DFN (T) & BERT & DeBERTa \\
\midrule
\multirow{5}{*}{\rotatebox{90}{Image}}
& CLIP (I)    & 0.220 & 0.128 & 0.218 & 0.203 & 0.207 \\
& SigLIP2 (I) & 0.274 & 0.153 & 0.278 & 0.249 & 0.256 \\
& DFN (I)     & 0.222 & 0.127 & 0.223 & 0.199 & 0.203 \\
& DinoV2      & 0.254 & 0.142 & 0.260 & 0.216 & 0.226 \\
& ViT         & 0.243 & 0.130 & 0.248 & 0.215 & 0.223 \\
& MambaVision & 0.219 & 0.117 & 0.223 & 0.190 & 0.197 \\
\midrule
\multirow{5}{*}{\rotatebox{90}{Text}}
& CLIP (T)    & 1     & 0.192 & 0.361 & 0.286 & 0.306 \\
& SigLIP2 (T) & 0.134 & 1     & 0.102 & 0.297 & 0.347 \\
& DFN (T)     & 0.345 & 0.173 & 1     & 0.282 & 0.301 \\
& BERT        & 0.237 & 0.172 & 0.238 & 1     & 0.275 \\
& DeBERTa     & 0.229 & 0.195 & 0.226 & 0.288 & 1     \\
\bottomrule
\end{tabular}
\end{table}

\begin{table}[tb]
\centering
\caption{$\textit{wMPPC}^{\textit{source}\to \textit{target}}$ (all layers) on COCO, for all 10 base models as source, image encoders as target}
\label{tab:wmppc-coco-base-all-10-img}
\begin{tabular}{@{}cl|ccccc@{}}
\toprule
\multicolumn{2}{c|}{\multirow{2}{*}{\diagbox[dir=NW]{Source}{Target}}}
    & \multicolumn{5}{c}{Image} \\
&   & CLIP (I) & SigLIP2 (I) & DFN (I) & DinoV2 & ViT \\
\midrule
\multirow{5}{*}{\rotatebox{90}{Image}}
& CLIP (I)    & 1     & 0.487 & 0.530 & 0.504 & 0.485 \\
& SigLIP2 (I) & 0.581 & 1     & 0.584 & 0.562 & 0.579 \\
& DFN (I)     & 0.527 & 0.500 & 1     & 0.522 & 0.485 \\
& DinoV2      & 0.499 & 0.499 & 0.523 & 1     & 0.481 \\
& ViT         & 0.459 & 0.458 & 0.465 & 0.455 & 1     \\
\midrule
\multirow{5}{*}{\rotatebox{90}{Text}}
& CLIP (T)    & 0.227 & 0.261 & 0.250 & 0.252 & 0.229 \\
& SigLIP2 (T) & 0.071 & 0.077 & 0.073 & 0.073 & 0.068 \\
& DFN (T)     & 0.240 & 0.281 & 0.269 & 0.270 & 0.251 \\
& BERT        & 0.154 & 0.171 & 0.167 & 0.168 & 0.152 \\
& DeBERTa     & 0.145 & 0.159 & 0.155 & 0.157 & 0.140 \\
\bottomrule
\end{tabular}

\end{table}

\begin{table}[tb]
\centering
\caption{$\textit{wMPPC}^{\textit{source}\to \textit{target}}$ (all layers) on COCO, for all 10 base models as source, text encoders as target}
\label{tab:wmppc-coco-base-all-10-txt}
\begin{tabular}{@{}cl|ccccc@{}}
\toprule
\multicolumn{2}{c|}{\multirow{2}{*}{\diagbox[dir=NW]{Source}{Target}}}
    & \multicolumn{5}{c}{Text} \\
&   & CLIP (T) & SigLIP2 (T) & DFN (T) & BERT & DeBERTa \\
\midrule
\multirow{5}{*}{\rotatebox{90}{Image}}
& CLIP (I)    & 0.240 & 0.150 & 0.241 & 0.213 & 0.207 \\
& SigLIP2 (I) & 0.273 & 0.174 & 0.275 & 0.241 & 0.237 \\
& DFN (I)     & 0.255 & 0.158 & 0.261 & 0.227 & 0.220 \\
& DinoV2      & 0.277 & 0.177 & 0.282 & 0.245 & 0.240 \\
& ViT         & 0.232 & 0.148 & 0.237 & 0.207 & 0.202 \\
\midrule
\multirow{5}{*}{\rotatebox{90}{Text}}
& CLIP (T)    & 1     & 0.343 & 0.508 & 0.408 & 0.399 \\
& SigLIP2 (T) & 0.363 & 1     & 0.361 & 0.479 & 0.433 \\
& DFN (T)     & 0.577 & 0.408 & 1     & 0.466 & 0.457 \\
& BERT        & 0.358 & 0.320 & 0.359 & 1     & 0.442 \\
& DeBERTa     & 0.323 & 0.330 & 0.324 & 0.437 & 1     \\
\bottomrule
\end{tabular}

\end{table}

\begin{table}[tb]
\centering
\caption{$\textit{wMPPC}^{\textit{source}\to \textit{target}}$ (last layer) on COCO, for all 10 base models as source, image encoders as target}
\label{tab:wmppc-coco-base-last-10-img}
\begin{tabular}{@{}cl|ccccc@{}}
\toprule
\multicolumn{2}{c|}{\multirow{2}{*}{\diagbox[dir=NW]{Source}{Target}}}
    & \multicolumn{5}{c}{Image} \\
&   & CLIP (I) & SigLIP2 (I) & DFN (I) & DinoV2 & ViT \\
\midrule
\multirow{5}{*}{\rotatebox{90}{Image}}
& CLIP (I)    & 1     & 0.353 & 0.335 & 0.266 & 0.282 \\
& SigLIP2 (I) & 0.352 & 1     & 0.356 & 0.278 & 0.304 \\
& DFN (I)     & 0.293 & 0.318 & 1     & 0.239 & 0.255 \\
& DinoV2      & 0.217 & 0.257 & 0.251 & 1     & 0.250 \\
& ViT         & 0.245 & 0.275 & 0.263 & 0.228 & 1     \\
\midrule
\multirow{5}{*}{\rotatebox{90}{Text}}
& CLIP (T)    & 0.224 & 0.260 & 0.247 & 0.208 & 0.227 \\
& SigLIP2 (T) & 0.084 & 0.095 & 0.089 & 0.075 & 0.080 \\
& DFN (T)     & 0.232 & 0.276 & 0.269 & 0.227 & 0.242 \\
& BERT        & 0.180 & 0.192 & 0.184 & 0.139 & 0.158 \\
& DeBERTa     & 0.156 & 0.164 & 0.152 & 0.112 & 0.130 \\
\bottomrule
\end{tabular}

\end{table}

\begin{table}[tb]
\centering
\caption{$\textit{wMPPC}^{\textit{source}\to \textit{target}}$ (last layer) on COCO, for all 10 base models as source, text encoders as target}
\label{tab:wmppc-coco-base-last-10-txt}
\begin{tabular}{@{}cl|ccccc@{}}
\toprule
\multicolumn{2}{c|}{\multirow{2}{*}{\diagbox[dir=NW]{Source}{Target}}}
    & \multicolumn{5}{c}{Text} \\
&   & CLIP (T) & SigLIP2 (T) & DFN (T) & BERT & DeBERTa \\
\midrule
\multirow{5}{*}{\rotatebox{90}{Image}}
& CLIP (I)    & 0.275 & 0.163 & 0.275 & 0.268 & 0.251 \\
& SigLIP2 (I) & 0.296 & 0.166 & 0.293 & 0.281 & 0.264 \\
& DFN (I)     & 0.240 & 0.138 & 0.242 & 0.223 & 0.210 \\
& DinoV2      & 0.235 & 0.132 & 0.238 & 0.195 & 0.178 \\
& ViT         & 0.232 & 0.127 & 0.230 & 0.209 & 0.195 \\
\midrule
\multirow{5}{*}{\rotatebox{90}{Text}}
& CLIP (T)    & 1     & 0.190 & 0.345 & 0.284 & 0.268 \\
& SigLIP2 (T) & 0.205 & 1     & 0.214 & 0.273 & 0.282 \\
& DFN (T)     & 0.419 & 0.237 & 1     & 0.342 & 0.324 \\
& BERT        & 0.244 & 0.193 & 0.271 & 1     & 0.338 \\
& DeBERTa     & 0.208 & 0.261 & 0.218 & 0.388 & 1     \\
\bottomrule
\end{tabular}

\end{table}

\begin{table}[tb]
\centering
\caption{$\textit{wMPPC}^{\textit{source}\to \textit{target}}$ (all layers) on Laion, for all 10 large models as source, image encoders as target}
\label{tab:wmppc-laion-large-all-10-img}
\begin{tabular}{@{}cl|ccccc@{}}
\toprule
\multicolumn{2}{c|}{\multirow{2}{*}{\diagbox[dir=NW]{Source}{Target}}}
    & \multicolumn{5}{c}{Image} \\
&   & CLIP (I) & SigLIP2 (I) & DFN (I) & DinoV2 & ViT \\
\midrule
\multirow{5}{*}{\rotatebox{90}{Image}}
& CLIP (I)    & 1     & 0.471 & 0.507 & 0.506 & 0.464 \\
& SigLIP2 (I) & 0.531 & 1     & 0.519 & 0.526 & 0.515 \\
& DFN (I)     & 0.428 & 0.379 & 1     & 0.409 & 0.365 \\
& DinoV2      & 0.566 & 0.531 & 0.551 & 1     & 0.532 \\
& ViT         & 0.401 & 0.394 & 0.387 & 0.409 & 1     \\
\midrule
\multirow{5}{*}{\rotatebox{90}{Text}}
& CLIP (T)    & 0.174 & 0.190 & 0.177 & 0.159 & 0.138 \\
& SigLIP2 (T) & 0.089 & 0.099 & 0.091 & 0.087 & 0.073 \\
& DFN (T)     & 0.194 & 0.212 & 0.196 & 0.174 & 0.152 \\
& BERT        & 0.148 & 0.152 & 0.140 & 0.133 & 0.113 \\
& DeBERTa     & 0.127 & 0.134 & 0.126 & 0.114 & 0.096 \\
\bottomrule
\end{tabular}

\end{table}

\begin{table}[tb]
\centering
\caption{$\textit{wMPPC}^{\textit{source}\to \textit{target}}$ (all layers) on Laion, for all 10 large models as source, text encoders as target}
\label{tab:wmppc-laion-large-all-10-txt}
\begin{tabular}{@{}cl|ccccc@{}}
\toprule
\multicolumn{2}{c|}{\multirow{2}{*}{\diagbox[dir=NW]{Source}{Target}}}
    & \multicolumn{5}{c}{Text} \\
&   & CLIP (T) & SigLIP2 (T) & DFN (T) & BERT & DeBERTa \\
\midrule
\multirow{5}{*}{\rotatebox{90}{Image}}
& CLIP (I)    & 0.162 & 0.072 & 0.186 & 0.151 & 0.136 \\
& SigLIP2 (I) & 0.215 & 0.093 & 0.253 & 0.201 & 0.181 \\
& DFN (I)     & 0.137 & 0.065 & 0.158 & 0.128 & 0.116 \\
& DinoV2      & 0.171 & 0.075 & 0.198 & 0.161 & 0.143 \\
& ViT         & 0.145 & 0.063 & 0.171 & 0.139 & 0.122 \\
\midrule
\multirow{5}{*}{\rotatebox{90}{Text}}
& CLIP (T)    & 1     & 0.582 & 0.663 & 0.583 & 0.547 \\
& SigLIP2 (T) & 0.621 & 1     & 0.613 & 0.696 & 0.732 \\
& DFN (T)     & 0.590 & 0.482 & 1     & 0.504 & 0.480 \\
& BERT        & 0.389 & 0.342 & 0.381 & 1     & 0.392 \\
& DeBERTa     & 0.446 & 0.482 & 0.435 & 0.520 & 1     \\
\bottomrule
\end{tabular}

\end{table}

\begin{table}[tb]
\centering
\caption{$\textit{wMPPC}^{\textit{source}\to \textit{target}}$ (last layer) on Laion, for all 10 large models as source, image encoders as target}
\label{tab:wmppc-laion-large-last-10-img}
\begin{tabular}{@{}cl|ccccc@{}}
\toprule
\multicolumn{2}{c|}{\multirow{2}{*}{\diagbox[dir=NW]{Source}{Target}}}
    & \multicolumn{5}{c}{Image} \\
&   & CLIP (I) & SigLIP2 (I) & DFN (I) & DinoV2 & ViT \\
\midrule
\multirow{5}{*}{\rotatebox{90}{Image}}
& CLIP (I)    & 1     & 0.228 & 0.215 & 0.129 & 0.172 \\
& SigLIP2 (I) & 0.252 & 1     & 0.244 & 0.150 & 0.212 \\
& DFN (I)     & 0.204 & 0.215 & 1     & 0.131 & 0.214 \\
& DinoV2      & 0.130 & 0.150 & 0.131 & 1     & 0.137 \\
& ViT         & 0.212 & 0.242 & 0.214 & 0.158 & 1     \\
\midrule
\multirow{5}{*}{\rotatebox{90}{Text}}
& CLIP (T)    & 0.148 & 0.153 & 0.139 & 0.084 & 0.114 \\
& SigLIP2 (T) & 0.088 & 0.092 & 0.083 & 0.042 & 0.058 \\
& DFN (T)     & 0.180 & 0.192 & 0.180 & 0.101 & 0.139 \\
& BERT        & 0.171 & 0.170 & 0.160 & 0.091 & 0.130 \\
& DeBERTa     & 0.130 & 0.138 & 0.123 & 0.068 & 0.090 \\
\bottomrule
\end{tabular}

\end{table}

\begin{table}[tb]
\centering
\caption{$\textit{wMPPC}^{\textit{source}\to \textit{target}}$ (last layer) on Laion, for all 10 large models as source, text encoders as target}
\label{tab:wmppc-laion-large-last-10-txt}
\begin{tabular}{@{}cl|ccccc@{}}
\toprule
\multicolumn{2}{c|}{\multirow{2}{*}{\diagbox[dir=NW]{Source}{Target}}}
    & \multicolumn{5}{c}{Text} \\
&   & CLIP (T) & SigLIP2 (T) & DFN (T) & BERT & DeBERTa \\
\midrule
\multirow{5}{*}{\rotatebox{90}{Image}}
& CLIP (I)    & 0.139 & 0.067 & 0.155 & 0.158 & 0.140 \\
& SigLIP2 (I) & 0.156 & 0.073 & 0.182 & 0.181 & 0.154 \\
& DFN (I)     & 0.123 & 0.061 & 0.145 & 0.135 & 0.117 \\
& DinoV2      & 0.086 & 0.040 & 0.098 & 0.093 & 0.083 \\
& ViT         & 0.133 & 0.058 & 0.141 & 0.145 & 0.124 \\
\midrule
\multirow{5}{*}{\rotatebox{90}{Text}}
& CLIP (T)    & 1     & 0.146 & 0.231 & 0.198 & 0.190 \\
& SigLIP2 (T) & 0.483 & 1     & 0.379 & 0.553 & 0.674 \\
& DFN (T)     & 0.238 & 0.147 & 1     & 0.217 & 0.215 \\
& BERT        & 0.221 & 0.166 & 0.221 & 1     & 0.256 \\
& DeBERTa     & 0.412 & 0.392 & 0.398 & 0.431 & 1     \\
\bottomrule
\end{tabular}

\end{table}

\begin{table}[tb]
\centering
\caption{$\textit{wMPPC}^{\textit{source}\to \textit{target}}$ (all layers) on Flowers-102, for all 10 large models as source, image encoders as target}
\label{tab:wmppc-flowers-large-all-10-img}
\begin{tabular}{@{}cl|ccccc@{}}
\toprule
\multicolumn{2}{c|}{\multirow{2}{*}{\diagbox[dir=NW]{Source}{Target}}}
    & \multicolumn{5}{c}{Image} \\
&   & CLIP (I) & SigLIP2 (I) & DFN (I) & DinoV2 & ViT \\
\midrule
\multirow{5}{*}{\rotatebox{90}{Image}}
& CLIP (I)    & 1     & 0.534 & 0.567 & 0.557 & 0.518 \\
& SigLIP2 (I) & 0.666 & 1     & 0.670 & 0.659 & 0.657 \\
& DFN (I)     & 0.519 & 0.479 & 1     & 0.493 & 0.452 \\
& DinoV2      & 0.598 & 0.575 & 0.591 & 1     & 0.575 \\
& ViT         & 0.457 & 0.466 & 0.456 & 0.480 & 1     \\
\midrule
\multirow{5}{*}{\rotatebox{90}{Text}}
& CLIP (T)    & 0.132 & 0.143 & 0.133 & 0.138 & 0.135 \\
& SigLIP2 (T) & 0.083 & 0.093 & 0.089 & 0.101 & 0.088 \\
& DFN (T)     & 0.161 & 0.170 & 0.162 & 0.167 & 0.164 \\
& BERT        & 0.121 & 0.147 & 0.134 & 0.146 & 0.133 \\
& DeBERTa     & 0.108 & 0.122 & 0.116 & 0.125 & 0.121 \\
\bottomrule
\end{tabular}

\end{table}

\begin{table}[tb]
\centering
\caption{$\textit{wMPPC}^{\textit{source}\to \textit{target}}$ (all layers) on Flowers-102, for all 10 large models as source, text encoders as target}
\label{tab:wmppc-flowers-large-all-10-txt}
\begin{tabular}{@{}cl|ccccc@{}}
\toprule
\multicolumn{2}{c|}{\multirow{2}{*}{\diagbox[dir=NW]{Source}{Target}}}
    & \multicolumn{5}{c}{Text} \\
&   & CLIP (T) & SigLIP2 (T) & DFN (T) & BERT & DeBERTa \\
\midrule
\multirow{5}{*}{\rotatebox{90}{Image}}
& CLIP (I)    & 0.240 & 0.095 & 0.259 & 0.154 & 0.134 \\
& SigLIP2 (I) & 0.238 & 0.100 & 0.256 & 0.157 & 0.141 \\
& DFN (I)     & 0.200 & 0.086 & 0.215 & 0.133 & 0.117 \\
& DinoV2      & 0.291 & 0.107 & 0.313 & 0.184 & 0.159 \\
& ViT         & 0.247 & 0.097 & 0.266 & 0.160 & 0.142 \\
\midrule
\multirow{5}{*}{\rotatebox{90}{Text}}
& CLIP (T)    & 1     & 0.440 & 0.630 & 0.516 & 0.488 \\
& SigLIP2 (T) & 0.508 & 1     & 0.503 & 0.462 & 0.465 \\
& DFN (T)     & 0.620 & 0.419 & 1     & 0.501 & 0.480 \\
& BERT        & 0.414 & 0.322 & 0.407 & 1     & 0.376 \\
& DeBERTa     & 0.348 & 0.343 & 0.348 & 0.351 & 1     \\
\bottomrule
\end{tabular}

\end{table}

\begin{table}[hb]
\centering
\caption{$\textit{wMPPC}^{\textit{source}\to \textit{target}}$ (last layer) on Flowers-102, for all 10 large models as source, image encoders as target}
\label{tab:wmppc-flowers-large-last-10-img}
\begin{tabular}{@{}cl|ccccc@{}}
\toprule
\multicolumn{2}{c|}{\multirow{2}{*}{\diagbox[dir=NW]{Source}{Target}}}
    & \multicolumn{5}{c}{Image} \\
&   & CLIP (I) & SigLIP2 (I) & DFN (I) & DinoV2 & ViT \\
\midrule
\multirow{5}{*}{\rotatebox{90}{Image}}
& CLIP (I)    & 1     & 0.418 & 0.410 & 0.317 & 0.335 \\
& SigLIP2 (I) & 0.413 & 1     & 0.434 & 0.383 & 0.370 \\
& DFN (I)     & 0.408 & 0.436 & 1     & 0.350 & 0.359 \\
& DinoV2      & 0.328 & 0.398 & 0.371 & 1     & 0.366 \\
& ViT         & 0.332 & 0.377 & 0.358 & 0.369 & 1     \\
\midrule
\multirow{5}{*}{\rotatebox{90}{Text}}
& CLIP (T)    & 0.238 & 0.287 & 0.260 & 0.175 & 0.239 \\
& SigLIP2 (T) & 0.079 & 0.087 & 0.085 & 0.097 & 0.080 \\
& DFN (T)     & 0.271 & 0.326 & 0.298 & 0.195 & 0.272 \\
& BERT        & 0.115 & 0.131 & 0.122 & 0.126 & 0.122 \\
& DeBERTa     & 0.105 & 0.111 & 0.114 & 0.113 & 0.113 \\
\bottomrule
\end{tabular}

\end{table}

\begin{table}[ht]
\centering
\caption{$\textit{wMPPC}^{\textit{source}\to \textit{target}}$ (last layer) on Flowers-102, for all 10 large models as source, text encoders as target}
\label{tab:wmppc-flowers-large-last-10-txt}
\begin{tabular}{@{}cl|ccccc@{}}
\toprule
\multicolumn{2}{c|}{\multirow{2}{*}{\diagbox[dir=NW]{Source}{Target}}}
    & \multicolumn{5}{c}{Text} \\
&   & CLIP (T) & SigLIP2 (T) & DFN (T) & BERT & DeBERTa \\
\midrule
\multirow{5}{*}{\rotatebox{90}{Image}}
& CLIP (I)    & 0.206 & 0.067 & 0.221 & 0.093 & 0.094 \\
& SigLIP2 (I) & 0.209 & 0.070 & 0.223 & 0.097 & 0.099 \\
& DFN (I)     & 0.223 & 0.070 & 0.238 & 0.099 & 0.102 \\
& DinoV2      & 0.150 & 0.067 & 0.160 & 0.086 & 0.084 \\
& ViT         & 0.202 & 0.066 & 0.216 & 0.092 & 0.094 \\
\midrule
\multirow{5}{*}{\rotatebox{90}{Text}}
& CLIP (T)    & 1     & 0.209 & 0.552 & 0.269 & 0.280 \\
& SigLIP2 (T) & 0.329 & 1     & 0.216 & 0.302 & 0.339 \\
& DFN (T)     & 0.582 & 0.201 & 1     & 0.273 & 0.281 \\
& BERT        & 0.224 & 0.214 & 0.203 & 1     & 0.240 \\
& DeBERTa     & 0.227 & 0.244 & 0.198 & 0.231 & 1     \\
\bottomrule
\end{tabular}

\end{table}

\clearpage
\section{Appendix: additional examples of visual features specific to VLMs}

\begin{figure}[ht]
    \centering
    \begin{subfigure}{0.45\textwidth}
        \centering
        \includegraphics[width=\linewidth]{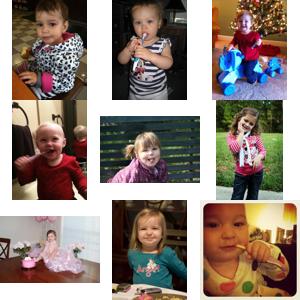}
        \caption{Age related}
    \end{subfigure}
    \hfill
    \begin{subfigure}{0.45\textwidth}
        \centering
        \includegraphics[width=\linewidth]{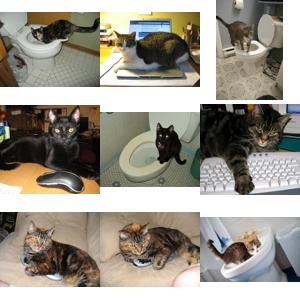}
        \caption{Pets with unusual behaviours}
    \end{subfigure}

    \vspace{0.5cm}

    \begin{subfigure}{0.45\textwidth}
        \centering
        \includegraphics[width=\linewidth]{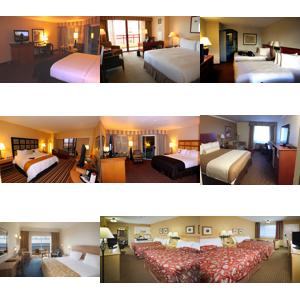}
        \caption{Rooms of the house (bedrooms)}
    \end{subfigure}
    \hfill
    \begin{subfigure}{0.45\textwidth}
        \centering
        \includegraphics[width=\linewidth]{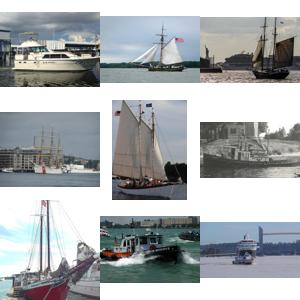}
        \caption{Vehicles (ships)}
    \end{subfigure}

    \vspace{0.5cm}

    \begin{subfigure}{0.45\textwidth}
        \centering
        \includegraphics[width=\linewidth]{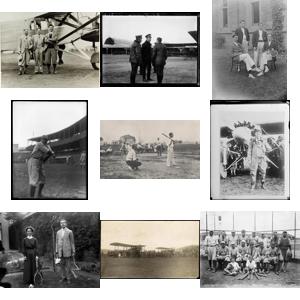}
        \caption{Old photos}
    \end{subfigure}
    \hfill
    \begin{subfigure}{0.45\textwidth}
        \centering
        \includegraphics[width=\linewidth]{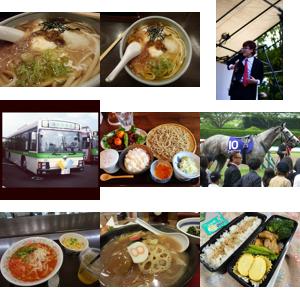}
        \caption{Geographical region (mostly via food)}
    \end{subfigure}

    \caption{Examples of visual features specific to VLMs mentioned in \autoref{subsec:visual-features-vlms}}
    \label{fig:examples-vlm-cs}
\end{figure}

\begin{figure}
    \centering
    \includegraphics[width=\linewidth]{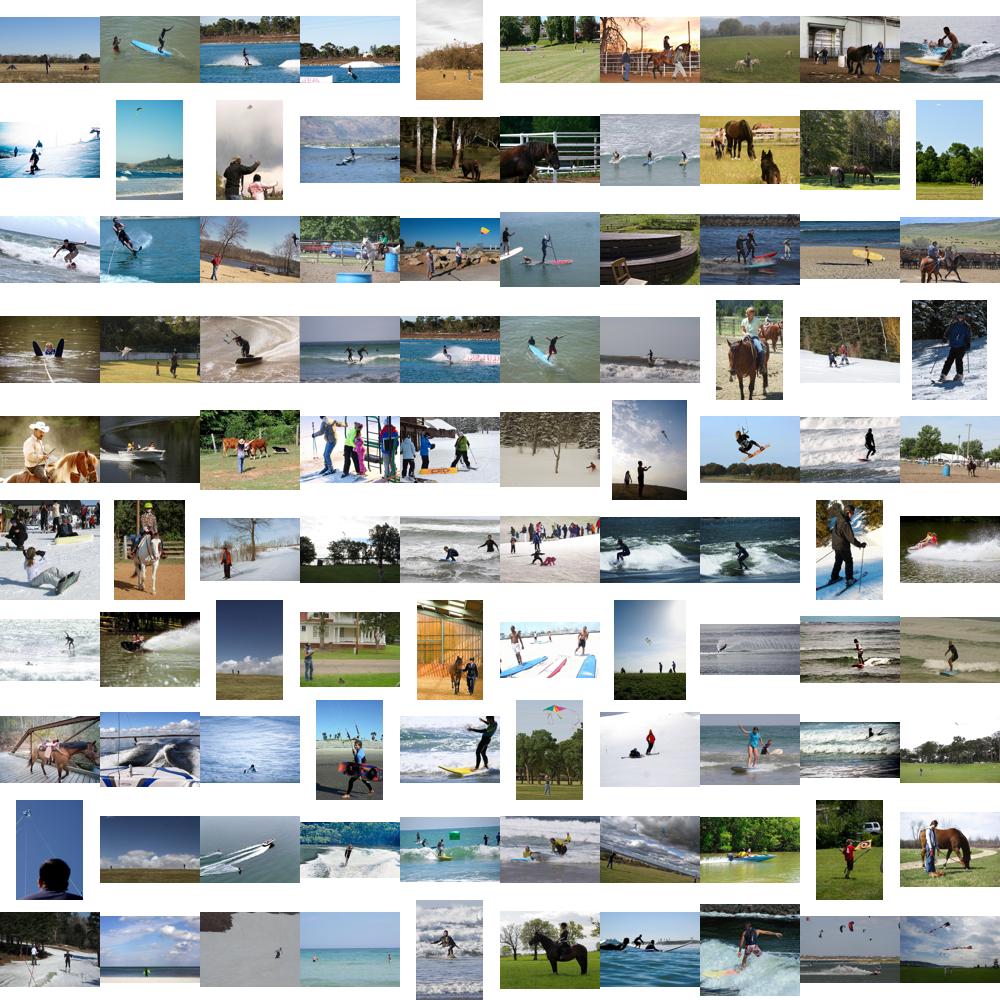}
    \caption{100 images corresponding to the highest activations of the feature associated to the verb ``to ride''}
    \label{fig:to_ride_feature}
\end{figure}

In \autoref{subsec:visual-features-vlms}, we provide a typology of features learnt on CLIP's visual encoder that are better share with other VLMs than with visual FMs. \autoref{fig:examples-vlm-cs} contains an example for each mentioned category. We display the feature corresponding to the highest Generalized Comparative Sharedness for each category, except for features present in \autoref{fig:cs-textual-examples}. In \autoref{fig:to_ride_feature}, we represent the 100 images corresponding to the highest activations of the feature associated to the verb ``to ride''. A larger number of examples is chosen here, in order to better represent the diversity of objects that activate this particular feature.

\clearpage
\section{Appendix: Code associated to the paper}\label{sec:app:code}
The code to reproduce the experiments is provided at \href{https://github.com/CEA-LIST/SAEshareConcepts}{https://github.com/CEA-LIST/SAEshareConcepts}. It is  developed from scratch and relies mainly on PyTorch~\cite{paszke2019pytorch} and numpy~\cite{harris2020numpy}. 

We used OpenCLIP~\cite{ilharco_gabriel2021openclip} and the Huggingface Transformers library~\cite{wolf2020transformers_lib} to handle models. As well, we relied on the Huggingface Datasets library~\cite{lhoest2021datasets_lib} to handle the datasets.

All these libraries are open source with permissive software license, as summarized in \autoref{tab:lib_licences}

\begin{table}[h]
    \centering
    \caption{Main libraries and code used in the paper}
    \label{tab:lib_licences}
    \begin{tabular}{ccc}
    \toprule
    Library & Source (URL) & Licence (URL) \\
    \midrule
    PyTorch & \href{https://github.com/pytorch/pytorch}{\faDownload} & \href{https://github.com/pytorch/pytorch/blob/main/LICENSE}{BSD}\\
    Numpy   & \href{https://github.com/numpy/numpy}{\faDownload} & \href{https://github.com/numpy/numpy/blob/main/LICENSE.txt}{BSD} \\
    OpenCLIP  & \href{https://github.com/mlfoundations/open_clip}{\faDownload} & \href{https://github.com/mlfoundations/open_clip/blob/main/LICENSE}{MIT}\\
    HF Transformers & \href{https://github.com/huggingface/transformers}{\faDownload} & \href{https://github.com/huggingface/transformers/blob/main/LICENSE}{Apache 2.0}\\
    HF Datasets & \href{https://github.com/huggingface/datasets}{\faDownload} & \href{https://github.com/huggingface/datasets/blob/main/LICENSE}{Apache 2.0}\\
    \bottomrule
    \end{tabular}
\end{table}

\end{document}